\documentclass[draft]{agujournal2019}
\usepackage{ragged2e}
\justifying
\AtBeginDocument{\justifying}
\usepackage{url} 
\usepackage{lineno}
\usepackage{natbib}
\usepackage[inline]{trackchanges} 
\usepackage{soul}
\usepackage{nicefrac}
\usepackage{booktabs}
\usepackage{amsmath}
\makeatletter
\renewcommand{\@oddhead}{}
\renewcommand{\@evenhead}{}
\makeatother
\draftfalse

\journalname{Journal of Advances in Modeling Earth Systems (JAMES)}

\begin{document}

\begin{center}
{\small\color{ltgray} This manuscript as presented here is a non-peer reviewed preprint submitted to \textit{JAMES}}
\end{center}

%
%


\title{Toward Mechanistic Interpretability of an \\ AI Foundation Model Fine-Tuned for \\ Atmospheric Chemistry}

%
%




\authors{Jason Y. Hu$^1$, Ivan Higuera-Mendieta$^2$, Patrick Obin Sturm$^3$, \\ and Makoto M. Kelp$^{2,4}$*}

\affiliation{1}{Stanford University, Department of Computer Science, Stanford, CA, USA.
}
\affiliation{2}{Stanford University, Department of Earth System Science, Stanford, CA, USA
}
\affiliation{3}{University of Southern California, Department of Earth Sciences, Los Angeles, CA, USA}
\affiliation{4}{University of  Utah, Department of Atmospheric Sciences, Salt Lake City, UT, USA}





\correspondingauthor{*Makoto M. Kelp}{makoto.kelp@utah.edu}



\begin{keypoints}
\item We present the first mechanistic interpretability analysis of an AI foundation model fine-tuned for atmospheric chemistry prediction.

\item Perturbation tests show Aurora captures some ozone-$\mathrm{NO}_x$ coupling but does not enforce the constraints of a process-based model.

\item  We identify and causally steer internal features to control chemical forecasts, opening a way toward interpreting model predictions.
\end{keypoints}

%
%

%
%


\begin{abstract}
Weather forecasting foundation models (FMs) are increasingly fine-tuned to predict air quality, offering fast global pollution forecasts at lower computational cost than conventional chemical transport models. These FMs are typically trained on reanalysis data and generate forecasts through autoregressive rollout. They do not explicitly represent governing physical or chemical processes. Therefore, high forecast skill does not reveal whether a model has learned physical mechanisms or exploits statistical regularities in its training data. Here, we present the first study of what a FM fine-tuned for atmospheric chemistry has learned by examining Microsoft's Aurora model. We impose controlled chemical perturbations on its forecasts and test them against known photochemical relationships. We then examine the internal representations that generate these forecasts. We find that Aurora captures a first-order ozone response to reactive nitrogen but does not enforce the chemical constraints that a process-based model encodes. It generates chemically inconsistent combinations of related species and relaxes localized emission features such as wildfire plumes toward background. Internally, its representations remain largely organized around the meteorology inherited during pretraining, with little structure specific to chemistry. Using sparse autoencoders, we identify internal components that causally control the chemical forecast but do not map cleanly onto individual atmospheric processes. This work provides a framework for testing whether AI forecasting systems learn atmospheric chemistry from reanalysis data. As these models are increasingly positioned to inform environmental policy decisions, we argue that composition forecasts should also be judged by their internal mechanisms rather than by benchmark skill alone.
\end{abstract}

\section*{Plain Language Summary}
Forecasting air pollution requires modeling atmospheric chemistry, which is computationally expensive for conventional numerical models. The rise of artificial intelligence (AI) foundation models has made it possible to adapt weather forecasting models to produce global air pollution forecasts faster and potentially more accurately. However, these AI models are trained to match large historical datasets rather than to follow the chemical laws that govern how pollutants form and react. As a result, it is unclear whether an accurate forecast implies that a model has learned atmospheric chemistry or instead relies on statistical patterns that may break down in future forecasts. We present the first study of the internal chemical reasoning of such a model by examining Microsoft’s Aurora model, adapted for air pollution forecasting. We find that Aurora reproduces some known chemistry, such as how reactive nitrogen can suppress ozone, but breaks basic chemical rules by predicting unrealistic concentrations and ignoring  features like wildfire smoke plumes. Internally, the model is organized mainly around weather patterns rather than chemistry. As these AI models are being considered for deployment in the real world, our work offers a way to judge whether they can be trusted based on how they represent chemistry internally.
\newpage

%
%

%


%
%
%
%

\section{Introduction}
Global modeling of atmospheric chemistry is critical to understanding how air pollution impacts human health, ecosystems, and climate forcing \citep{brasseurModelingAtmosphericChemistry2017}. Chronic exposure to elevated pollutant concentrations is associated with asthma, cardiovascular and respiratory disease, and premature mortality \citep{anenbergEstimatesGlobalBurden2018a, qiuWildfireSmokeExposure2025b}, while reactive trace gases and aerosols further impair crop productivity, ecosystem function, and climate feedbacks \citep{millsOzonePollutionWill2018, weberChemistrydrivenChangesStrongly2022, zhuUncertainNaturalEmissions2026}. Thus, accurate forecasts of atmospheric composition and long-range transport can support both public health interventions and regulatory decision making \citep{kellerDescriptionNASAGEOS2021}. Yet atmospheric chemistry is the most computationally demanding component of Earth system and climate models, owing to the large number of coupled chemical species and their interactions with transport across all scales \citep{easthamGEOSChemHighPerformance2018}. The US National Research Council identified atmospheric chemistry as a priority frontier for Earth system model development more than a decade ago, but incorporating detailed chemistry into global models remains limited by its substantial computational cost \citep{nationalresearchcouncilNationalStrategyAdvancing2012}.

Artificial intelligence (AI) has emerged as a promising approach to reduce this computational burden, particularly by emulating the most computationally expensive components of chemical transport models (CTMs) \citep{hickmanApplicationsMachineLearning2025a}. Previous studies show that neural networks can approximate chemical mechanisms at far lower cost \citep{kelpStableGeneralMachineLearned2020}, but they also reveal a central challenge of controlling error growth during long autoregressive integrations \citep{kelpOrdersofmagnitudeSpeedupAtmospheric2018, kellerApplicationRandomForest2019}. More recent studies have deployed such emulators online within CTMs. \cite{kelpOnlineLearnedNeuralNetwork2022}  implemented a neural network chemical solver in the GEOS-Chem CTM and achieved stable one-year simulations at reduced cost, and \cite{xiaAdvancingSophisticatedPhotochemistry2025} coupled a transformer-based solver into WRF-Chem, demonstrating large speedups for short regional simulations over East Asia. Parallel efforts seek to embed physical constraints into these models, including mass- and stoichiometry-conserving guardrails \citep{sturmConservationLawsNeural2022a, sturmMassEnergyconservingFramework2020a}, guided mechanism reduction \citep{shenAdaptiveMethodSpeeding2020, wiserAMOREIsopreneV10New2023}, and learned compression of chemical tracers during transport \citep{sturmAdvectingSuperspeciesEfficiently2023}. Despite this progress, most AI applications in atmospheric chemistry remain offline or diagnostic, including bias corrections \citep{ivattImprovingPredictionAtmospheric2020b, liuCorrectingOzoneBiases2022a, oakBiascorrectedGEMSGeostationary2024}, statistical downscaling \citep{yuDeepLearningbasedDownscaling2021, xiaoGeneratingLongterm200320202022, gouldsbroughMachineLearningApproach2024a}, or building reanalysis-like datasets \citep{diEnsemblebasedModelPM252019,shenEnhancingGlobalEstimation2024, liGrowingImpactsFire2026}. Comparatively few are dynamically coupled within conventional CTM modeling systems, in part because online integration remains prone to coupled instabilities \citep{kelpOrdersofmagnitudeSpeedupAtmospheric2018, brenowitzInterpretingStabilizingMachineLearning2020, raspCoupledOnlineLearning2020a}.  

Foundation models (FMs) for weather and climate extend AI emulation from individual model components to the full atmospheric state, learning to forecast weather and climate end-to-end from large Earth system datasets. Because FMs predict the entire state directly, they avoid embedding an emulated model component inside a CTM where it can suffer instability or error growth during online integration. A single FM can also be pretrained once and then fine-tuned for many downstream tasks at a small fraction of the cost of building a dedicated numerical model. This efficiency is expected to make fine-tuned FMs an increasingly widespread tool for Earth system prediction \citep{laiMachineLearningClimate2024, hickmanApplicationsMachineLearning2025a}. Recent weather FMs now outperform traditional numerical weather prediction systems \citep{biAccurateMediumrangeGlobal2023, lamLearningSkillfulMediumrange2023, nguyenClimaXFoundationModel2023, bodnarFoundationModelEarth2025}, motivating public agencies and private companies to develop and deploy their own AI models \citep{langAIFSECMWFsDatadriven2024a, sadeghitabasGFSPoweredMachineLearning2025}. However, as these AI systems move toward operational use in settings that inform public health and policy, their training data, fine-tuning procedures, and evaluation protocols are not always fully open or reproducible \citep{camps-vallsArtificialIntelligenceModeling2025}. FMs are also typically trained on reanalysis data and generate forecasts through autoregressive rollout rather than by explicitly representing governing physical or chemical processes. As such, high forecast skill does not reveal whether a model has learned physical mechanisms or is exploiting statistical regularities in its training data \citep{corleyNoOneKnows2026,fengBenchmarkingImprovingMonitors2026}. Microsoft's Aurora model extends this paradigm to atmospheric chemistry by fine-tuning a pretrained Earth system FM to forecast air pollution \citep{bodnarFoundationModelEarth2025}. This Aurora model produces composition forecasts without explicit chemical mechanisms, process-level constraints, or time-varying emissions during rollout. It therefore raises a central question for AI atmospheric chemistry: can an end-to-end FM learn chemistry from reanalysis data alone?


Mechanistic interpretability offers a new approach for evaluating AI atmospheric chemistry models by probing how air pollution forecasts are produced through the model's internal `circuitry'. Existing explainable AI (XAI) methods in climate science are primarily designed to attribute predictions to input variables and to evaluate the reliability of those attribution maps \citep{lundbergUnifiedApproachInterpreting2017, tomsPhysicallyInterpretableNeural2020}. These methods are useful for asking which parts of an atmospheric state a forecast is sensitive to, but they are not sufficient for determining whether the model constructs forecasts through internal representations of coupled chemical processes \citep{silvaLimitationsXAIMethods2024, bommerFindingRightXAI2024}. Mechanistic interpretability, developed originally for large language models, instead examines model internals (e.g., activations, residual-stream representations) to identify which intermediate quantities contribute to a prediction \citep{olahZoomIntroductionCircuits2020, nandaProgressMeasuresGrokking2023, bereskaMechanisticInterpretabilityAI2024}. These internal states can be visualized across space and time, compared across forecast initializations, or perturbed in steering experiments to quantify their causal effect on outputs \citep{chalnevImprovingSteeringVectors2024, chouCausalLanguageControl2025}. The approach is well suited to atmospheric chemistry where many behaviors emerge from relationships among chemical species, emissions, and transport rather than from any single predicted field. Yet mechanistic interpretability techniques have rarely been applied to weather or Earth system models \citep{macmillanMechanisticUnderstandingDatadriven2025a, kasteleynPhysMetricsWeatherEvaluationFramework2026} and never to AI atmospheric chemistry modeling in particular. 


Here, we present the first mechanistic interpretability analysis of an AI FM fine-tuned for atmospheric chemistry. We study Aurora's air pollution model and probe whether its forecasts are supported by chemically-informed internal mechanisms. To assess chemical behavior, we examine whether Aurora preserves relationships that emerge from processes it was never explicitly trained to represent. These include reactive-nitrogen partitioning, inferred photolysis rates, ozone responses to nitrogen oxide ($\mathrm{NO}_x$) perturbations, and wildfire emission and transport. To probe the model internally, we test whether its latent representations organize around chemical processes or primarily reflect the meteorological structure inherited from pretraining. Finally, we train and release \textit{AuroraScope}, a suite of sparse autoencoders (SAEs) for Aurora. We use feature visualization and steering experiments to test whether learned internal features causally influence chemical forecasts. These analyses establish a framework for evaluating whether end-to-end AI forecasting systems learn the principles of atmospheric chemistry from data alone.


\section{Materials and Methods}

We organize the following sections to move from diagnosing chemical forecast behavior to internal mechanisms in Aurora. First, we test whether Aurora's predicted chemical fields preserve expected atmospheric chemistry relationships using diagnostics of photochemical species coupling, controlled $\mathrm{NO}_x$ perturbations, and wildfire plume persistence. Second, we analyze encoder and decoder latent representations to determine whether Aurora's broad internal spatial structure reflects atmospheric chemistry or is primarily meteorological. Finally, we train SAEs on the model operator's activations and use feature visualization and steering experiments to test whether individual learned features causally influence chemical forecasts.

\subsection{Aurora Foundation Model Fine-Tuned for Air Pollution}   

We analyze Aurora Air Pollution, the Aurora FM checkpoint fine-tuned for global atmospheric chemistry forecasting (available at microsoft/aurora/aurora-0.4-air-pollution on Hugging Face). Aurora is a 1.3 billion parameter Earth system FM pretrained on more than one million hours of heterogeneous atmospheric data, including ERA5 reanalysis, CMIP6 climate simulations, GFS operational analyses, and CAMS atmospheric chemistry data. For air pollution forecasting, Bodnar et al. fine-tuned Aurora on CAMS analysis data from October 2017 to May 2022 and also incorporated lower-resolution CAMS reanalysis data from January 2003 to December 2021. They describe Aurora fine-tuning as a two-stage procedure consisting of short lead time fine-tuning of the pretrained weights followed by long lead time rollout fine-tuning using Low-Rank Adaptation (LoRA) \citep{huLoRALowRankAdaptation2021}. The resulting model predicts 12-h changes in atmospheric composition at 0.4 degree horizontal resolution during autoregressive rollout. The model can generate five-day global air pollution forecasts in under a minute. In Bodnar et al.'s evaluation across June 2022 to November 2022, the model is within 20\% root mean square error (RMSE) of CAMS on 96\% of targets and matches or outperforms CAMS on 76\% of targets.

Aurora Air Pollution predicts trace gas concentrations of carbon monoxide ($\mathrm{CO}$), nitric oxide (NO), nitrogen dioxide ($\mathrm{NO}_2$), ozone ($\mathrm{O}_3$), and sulfur dioxide ($\mathrm{SO}_2$), along with particulate matter aerosol and total-column composition fields. Several of these variables are directly related to regulated air pollutants, including criteria pollutants under the U.S. National Ambient Air Quality Standards. Other species provide information about combustion, oxidation chemistry, and long range transport. The model also uses static pollution-related surface fields, including monthly average anthropogenic emissions for ammonia, $\mathrm{CO}$, $\mathrm{NO}_x$, and $\mathrm{SO}_2$, which provide fixed spatial information about typical emissions patterns. These static fields help constrain the model geographically but do not provide time-varying emissions. Although Aurora uses pollutant-specific transformations and clipping to stabilize forecasts, it does not enforce positivity among species \citep{bodnarFoundationModelEarth2025}.

Aurora Air Pollution’s architecture determines both the spatial resolution of its forecasts and the internal activations available for mechanistic analysis. The model forecasts the next atmospheric state from two preceding atmospheric states separated by 12-h. It consists of three main components (Figure 1). An encoder divides each gridded field into small spatial patches and embeds each patch as a ``token" (i.e., a vector representation of that patch). A U-Net operator propagates information across tokens. A decoder maps the processed tokens back to gridded atmospheric variables. Every atmospheric state contains surface variables and three-dimensional atmospheric variables on a global 0.4 degree latitude-longitude grid. Each surface variable is represented as a $451 \times 900$ array, while each atmospheric variable is represented as a $13 \times 451 \times 900$ array corresponding to 13 pressure levels ranging from 1000 hPa near the surface to 50 hPa in the upper stratosphere.

\begin{figure}
    \centering
    \includegraphics[width=1\linewidth]{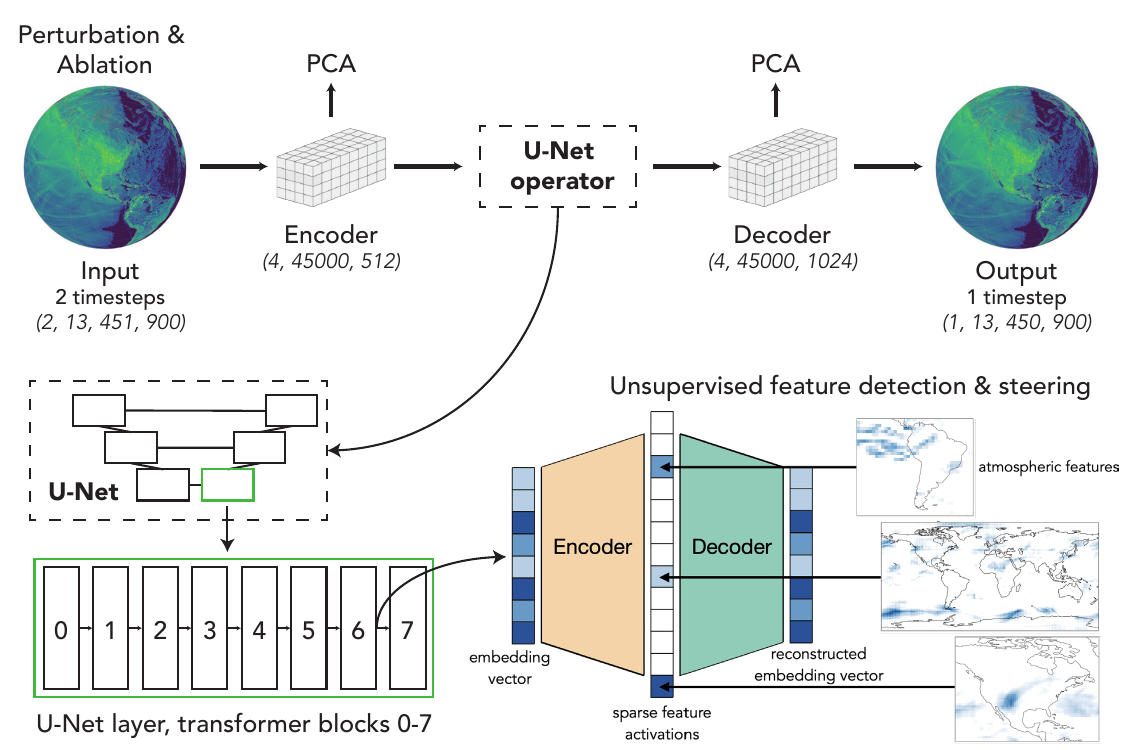}
    \caption{\textbf{Mechanistic interpretability of the Aurora Earth system foundation model.} For each atmospheric chemical species, Aurora ingests two input timesteps at 13 pressure levels from 1000 hPa to 50 hPa across 451 latitude and 900 longitude cells. All atmospheric chemistry predictions occur on 12-h time steps. The model encodes all atmospheric variables into 4 latent levels with $3\times3$ patching of the latitude and longitude grid. Each patch is embedded into 512 dimensions. All patches are fed through the U-Net operator, which returns a 1024-dimensional vector. The U-Net operator has 3 encoder and 3 decoder stages, with each stage consisting of multiple 3D Swin transformer blocks. To probe the atmospheric behavior of Aurora, we: (1) perturb input concentrations and analyze the effects on output concentrations, (2) analyze underlying patterns found in the encoder and decoder latent levels using principal components analysis (PCA), and (3) train sparse autoencoders on the residual stream of each operator transformer block to discover sparse feature activations.}
    \label{fig:overview}
\end{figure}

Because our analyses probe both encoded latent states and operator activations, we briefly summarize how gridded fields are transformed inside Aurora. The encoder groups each $451 \times 900$ horizontal grid into $3\times3$ patches, producing a $150 \times 300$ grid of patch tokens after truncating the final latitude row. For each patch, the encoder compresses all surface variables into a 512-dimensional vector producing a $150 \times 300 \times 512$ surface latent level. The 13 atmospheric pressure levels are compressed into three additional latent levels yielding four input latent levels in total. The operator processes these input latent levels through a U-Net with three encoder stages and three decoder stages connected by skip connections. The operator output has dimensions $150 \times 300 \times 1024$ with the doubled embedding dimension resulting from the concatenation skip connection between the first encoder stage and final decoder stage. 

Throughout this study, CAMS fields serve two roles. CAMS composition fields provide the initial conditions for Aurora Air Pollution forecasts, while the CAMS forecasts initialized at the same times provide reference simulations for comparison. Unless otherwise specified, comparisons between Aurora and CAMS are made at matched initialization times and forecast lead times. We extend the initial validation period by evaluating forecasts initialized from June 1, 2022 at 00:00 UTC through May 31, 2023 at 12:00 UTC. For selected experiments, we use the September 1, 2022 at 00:00 UTC CAMS initialization time step analyzed by Bodnar et al.

\subsection{Diagnosing Atmospheric Chemistry Behavior} 

\subsubsection{Photochemistry and Chemical Species Coupling}

Aurora Air Pollution forecasts a limited set of chemically-active trace gases, but it does not explicitly represent intermediate species, photolysis rates, or kinetic mechanisms that govern their evolution. We test whether the model preserves two relationships among the species it predicts. Both diagnostics are derived only from $\mathrm{NO}$, $\mathrm{NO}_2$, and ozone, which are linked through fast photochemical cycling.

The first diagnostic is the fraction of represented $\mathrm{NO}_x$ present as $\mathrm{NO}_2$, defined as:

\begin{equation}
f_{\mathrm{NO}_2} = \frac{[\mathrm{NO}_2]}{[\mathrm{NO}_2] + [\mathrm{NO}]}    
\label{eq:frac_no}
\end{equation}

This ratio measures how the model partitions reactive nitrogen between $\mathrm{NO}$ and $\mathrm{NO}_2$. In the atmosphere, $\mathrm{NO}_x$ partitioning exhibits strong diurnal and chemical structure because sunlight photolyzes $\mathrm{NO}_2$ to produce $\mathrm{NO}$, while $\mathrm{NO}$ reacts with ozone to regenerate $\mathrm{NO}_2$  \citep{leightonPhotochemistryAirPollution2012}. Although Aurora is not explicitly provided with photolysis rates or intermediate radicals, its predicted $\mathrm{NO}$ and $\mathrm{NO}_2$ fields should still maintain physically plausible partitioning if the model has learned chemical relationships between these species.

The second diagnostic is an idealized Leighton-derived $\mathrm{NO}_2$ photolysis rate \citep{leightonPhotochemistryAirPollution2012} computed from the predicted $\mathrm{NO}$, $\mathrm{NO}_2$, and ozone fields. Under a photostationary assumption with Leighton ratio equal to one, the $\mathrm{NO}_2$ photolysis rate $J_1$ can be approximated as:

\begin{equation}
J_1 \approx \frac{k_3\,[\mathrm{NO}][\mathrm{O}_3]}{[\mathrm{NO}_2]}    
\label{eq:leighton}
\end{equation}

where $k_3$ is the rate constant for the reaction $\mathrm{NO} + \mathrm{O}_3 \rightarrow \mathrm{NO}_2 + \mathrm{O}_2$. We do not interpret this quantity as a globally valid estimate of the true photolysis rate, because the photostationary assumption can break down under transport, fresh emissions, and nighttime settings, among other conditions \citep{calvertAcidGenerationTroposphere1983}. Instead, we use it as an idealized test of consistency for the subset of fast $\mathrm{NO}_x$-ozone chemistry represented in the model. Because this diagnostic contains $\mathrm{NO}_2$ in the denominator and is only physical for positive concentrations, we mask grid cells where $\mathrm{NO}$, $\mathrm{NO}_2$, or ozone is nonpositive before interpreting the Leighton-derived value. We retain the spatial extent of these masked regions as a separate indicator of nonphysical species predictions rather than interpreting negative or undefined values as physical photolysis rates.

For both Aurora Air Pollution and CAMS, we compute the $\mathrm{NO}_x$ partitioning ratio and the Leighton-derived relationship at the 12-h forecast step and compare their spatial distributions.

\subsubsection{$\mathrm{NO}_x$ Perturbation Experiments}

We perturb the $\mathrm{NO}$ and $\mathrm{NO}_2$ initial conditions in Aurora Air Pollution to test whether the model has learned chemically plausible ozone responses to changes in reactive nitrogen. In tropospheric chemistry, $\mathrm{NO}_x$ both contributes to ozone production and can suppress ozone through $\mathrm{NO}$ titration, producing a nonlinear and regime-dependent relationship between $\mathrm{NO}_x$ and ozone \citep{seinfeldAtmosphericChemistryPhysics2016}. The sign and magnitude of the ozone response depend on chemical environment, including the availability of volatile organic compounds ($\mathrm{VOCs}$) and radicals that control the conversion of $\mathrm{NO}$ to $\mathrm{NO}_2$. Although the model does not include $\mathrm{VOCs}$ or radicals as prognostic variables, it may encode spatially varying chemical regimes implicitly through correlations in the training data (for example, biogenically influenced regions versus urban areas). 

We test this behavior using controlled $\mathrm{NO}_x$ perturbation experiments. For each initialization, we first convert the $\mathrm{NO}$ and $\mathrm{NO}_2$ input fields to mixing ratio units. We add a constant offset to both species across all grid cells, pressure levels, and input timesteps. We then run a five-day rollout consisting of ten 12-h forecast steps. We apply three perturbation magnitudes: +0.5 ppb (representing a small perturbation comparable to background variability), +5 ppb (representing a pollution-scale perturbation), and +20 ppb (representing an extreme disturbance test). We repeat each perturbation experiment for all CAMS initializations from June 1, 2022 at 00:00 UTC through May 31, 2023 at 12:00 UTC. We compare the perturbed rollouts against the unperturbed rollout to evaluate the resulting ozone response.

\subsubsection{Wildfire Persistence}

We test whether Aurora Air Pollution can maintain emissions-driven pollution anomalies using wildfire smoke plumes as case studies. Wildfires emit concentrated pulses of $\mathrm{CO}$, $\mathrm{NO}_x$, aerosols, and other pollutants that produce localized composition anomalies and downwind transport features. In these experiments, wildfire information can enter Aurora only through the initialized atmospheric composition fields, which are derived from CAMS fields containing fire-related composition anomalies. These experiments test whether Aurora can maintain and transport an existing wildfire plume signal during rollout, despite lacking explicit time-varying fire emissions or process-based emissions tendencies during the forecast.

We select three wildfire events from the analysis period based on the largest total column $\mathrm{CO}$ enhancements in CAMS. The selected cases are in Spain (initialized on July 15, 2022 at 12:00 UTC), Alaska (initialized on July 3, 2022 at 00:00 UTC), and Central Russia (initialized on August 19, 2022 at 12:00 UTC). For each event, we compare Aurora Air Pollution forecasts with CAMS forecasts initialized at the same time and evaluate the persistence and transport of total column $\mathrm{CO}$ over the forecast rollout.

\subsection{Latent Representation Qualitative Analysis}

Aurora Air Pollution's latent representations help illustrate whether the model's internal spatial structure reflects atmospheric chemistry patterns or primarily meteorological organization. To visualize these representations, we apply principal component analysis (PCA) to the patch-token embeddings at each of the four input latent levels (encoder) and four output latent levels (decoder). For a single initialization, the PCA input matrix therefore has one row for each spatial patch token and one column for each embedding channel. For monthly and annual analyses, we concatenate patch-token embeddings across the selected forecast initializations before fitting the PCA. The embedding dimension is 512 for input latent levels and 1024 for output latent levels.

We extract the first three principal components and map the projected component scores to RGB channels with per-channel normalization. This produces spatial visualizations of the dominant directions of latent variation. For the larger monthly and annual datasets, we use incremental PCA as a memory-efficient implementation \citep{gomez-pedreroIncrementalPCAAlgorithm2022}. We perform the analysis across three temporal subsets: a single initialization on September 1, 2022 at 00:00 UTC; all initializations from September 2022; and all initializations in the full year-long analysis period. Comparing these subsets allows us to distinguish spatial patterns that are specific to one forecast initialization from persistent structures that appear across seasonal and annual variability. Because Aurora compresses the surface variables into one latent level (latent level 0) and the 13 atmospheric pressure levels into three additional latent levels (latent levels 1, 2, 3), we also examine whether latent-level structure varies with the vertical information encoded at each latent level.

\subsection{Sparse Feature Discovery \& Steering}

We train a suite of SAEs on Aurora Air Pollution's operator activations and release them as \textit{AuroraScope}. SAEs decompose dense model activations into sparse learned features, which allow us to identify localized or structured activation patterns that may not be visible in the full residual stream \citep{cunninghamSparseAutoencodersFind2023}. We train one SAE for each of the 48 transformer blocks in the model's operator. 

Each SAE is trained with an $8\times $ expansion factor on activations from rollouts across the full analysis period. For each operator layer, training examples are sampled from residual stream activations across forecast initializations and rollout steps in the full analysis period with each example corresponding to one patch-token activation vector. We subsample patch tokens across space, time, and forecast lead time to avoid training only on geographically or temporally adjacent activations. Models are trained for 3 epochs using Adam \citep{kingmaAdamMethodStochastic2017} with a learning rate of $4\times 10^{-4}$. Training required approximately 400 A100 GPU-hour equivalents. We evaluate reconstruction quality using fraction of variance explained, defined as $\mathrm{FVE} = 1 - \nicefrac{MSE}{\text{Var}(\mathbf{x})}$.

We interpret each \textit{AuroraScope} feature by visualizing its per-token variance contribution. For feature $k$ at operator block $b$, we define the contribution at token t as in Equation \ref{eq:kv_c}:
\begin{equation}
    v_k(t) = \frac{h_k(t)^2}{\|\mathbf{x}(t) - \mathbf{b}_{\mathrm{dec}}\|_2^2}
\label{eq:kv_c}
\end{equation}

where $h_k(t)$  is the SAE activation of feature $k$ on patch token $t$, $\mathbf{x}(t)$ is the residual stream vector for patch token $t$, and $\mathbf{b}_{\mathrm{dec}}$ is the decoder bias. Because the decoder columns are unit-normalized, $v_k(t)$ is bounded in $[0,1]$ and represents the fraction of token reconstruction variance attributable to feature $k$ alone. This normalization allows feature activations to be compared across operator layers with different dimensionality.

We also perform steering interventions to test whether selected sparse features causally influence chemical forecasts. For a target feature $k$ at operator block $b$, we zero-ablate the feature globally by removing its SAE decoder contribution from the residual stream at every patch token. We then complete the rollout and compare the resulting forecast with the unperturbed forecast at 12-h lead time. To select features for steering, we rank all features across all 48 operator layers by maximum per-token variance contribution and select the global top ten. From these ten ablations, we select two steering case studies (Section 3.5).

\section{Results and Discussion}

\subsection{Photochemistry and Species Coupling}

We first evaluate whether Aurora Air Pollution preserves relationships among chemically coupled species during a forecast. Figure 2 shows that in CAMS, the $\mathrm{NO}_x$ partitioning ratio contains coherent large-scale spatial structure with much of the global surface field close to $\mathrm{NO}_2$-dominated conditions. This ratio is lower across sunlit regions where $\mathrm{NO}_2$ photolysis shifts reactive nitrogen toward $\mathrm{NO}$. Aurora reproduces some broad geographic structure, but its $\mathrm{NO}_x$ partitioning field is much noisier and contains large masked regions where at least one component concentration is nonpositive. The Aurora field also shows visible $3 \times3$ patch-scale artifacts. These masked regions indicate species combinations that cannot support diagnostics requiring positive concentrations.

The Leighton-derived relationship shows a similar pattern. CAMS produces a spatially coherent field with regional structure consistent with $\mathrm{NO}_2$ photochemistry, while Aurora produces a noisier field with visible artifacts, ship track structures, and extensive masking caused by nonpositive species concentrations. As this diagnostic relies on an idealized photostationary assumption, we do not interpret the inferred value as a globally valid photolysis rate. Instead, the result indicates that Aurora's predicted $\mathrm{NO}$, $\mathrm{NO}_2$, and ozone fields are often not mutually consistent under a simplified fast-chemistry relationship. These results suggest that Aurora Air Pollution can reproduce some broad cross-species spatial structure at short lead time, but also generates nonphysical concentrations and chemically-inconsistent species combinations that are not constrained by an explicit chemical mechanism.

\begin{figure}
    \centering
    \includegraphics[width=1\linewidth]{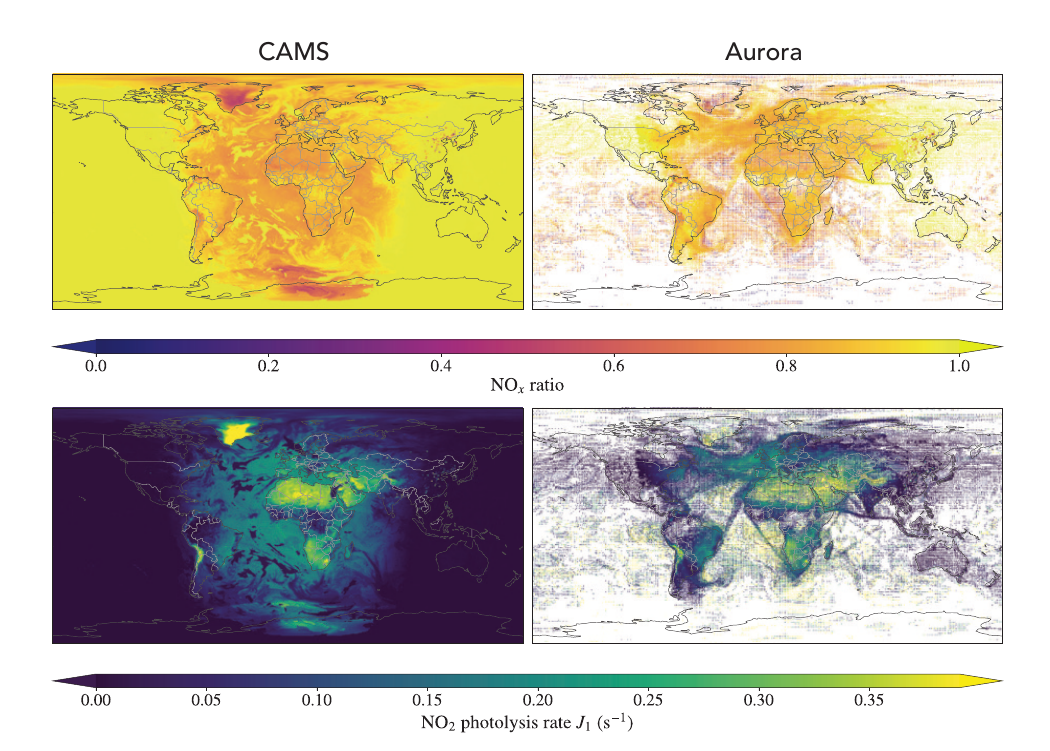}
    \caption{\textbf{Calculated $\mathrm{NO}_x$ ratio and $\mathrm{NO}_2$ photolysis rate in CAMS and Aurora.} Both simulations are on September 1 2022 at 12:00 UTC. The $\mathrm{NO}_x$ ratio is calculated as $\frac{[NO_2]}{[NO]+[NO_2]}$. The $\mathrm{NO}_2$ photolysis rate is calculated as $\frac{k_3[NO][O_3]}{[NO_2]}$, which assumes photostationarity based on the Leighton relationship. Depicted is the first rollout step (+12h) for both models at surface level (1000 hPa). White areas on the map indicate where the model predicts a negative concentration for at least one component species within the calculation.}
    \label{fig:NOx_leighton}
\end{figure}

\subsection{$\mathrm{NO}_x$ Perturbation Response}

Controlled $\mathrm{NO}_x$ perturbations reveal that Aurora Air Pollution has learned an ozone response to reactive nitrogen, but the response is not uniformly chemically interpretable (Figure 3). At 12-h lead time, the +5 ppb $\mathrm{NO}_x$ perturbation produces widespread ozone decreases relative to the baseline forecast, consistent with rapid ozone suppression by enhanced $\mathrm{NO}$ in many regions. In the +5 ppb $\mathrm{NO}_x$ experiment, the median surface ozone concentration is 6.1 ppb lower than in the baseline forecast and ozone is reduced in 89\% of surface grid cells. The perturbed surface ozone distribution is displaced toward lower values and extends into nonphysical negative ozone concentrations. This behavior suggests that Aurora has learned a strong ozone-suppression response to $\mathrm{NO}_x$, but without the positivity constraints that would be imposed by a process-based chemical model. Aurora is trained to minimize aggregate forecast error without explicit physical constraints. As a result, it can tolerate locally impossible values such as negative concentrations. Negative ozone is simply the most readily detected of such nonphysical violations. The strictly positive but chemically inconsistent species combinations in our photochemical diagnostics (Section 3.1) suggest that comparable inconsistencies also arise among physically admissible values, where they may be much harder to detect. The +20 ppb experiment produces a qualitatively similar but stronger response with ozone suppressed across nearly all regions, consistent with widespread titration. (Figure S1).

\begin{figure}
    \centering
    \includegraphics[width=.95\linewidth]{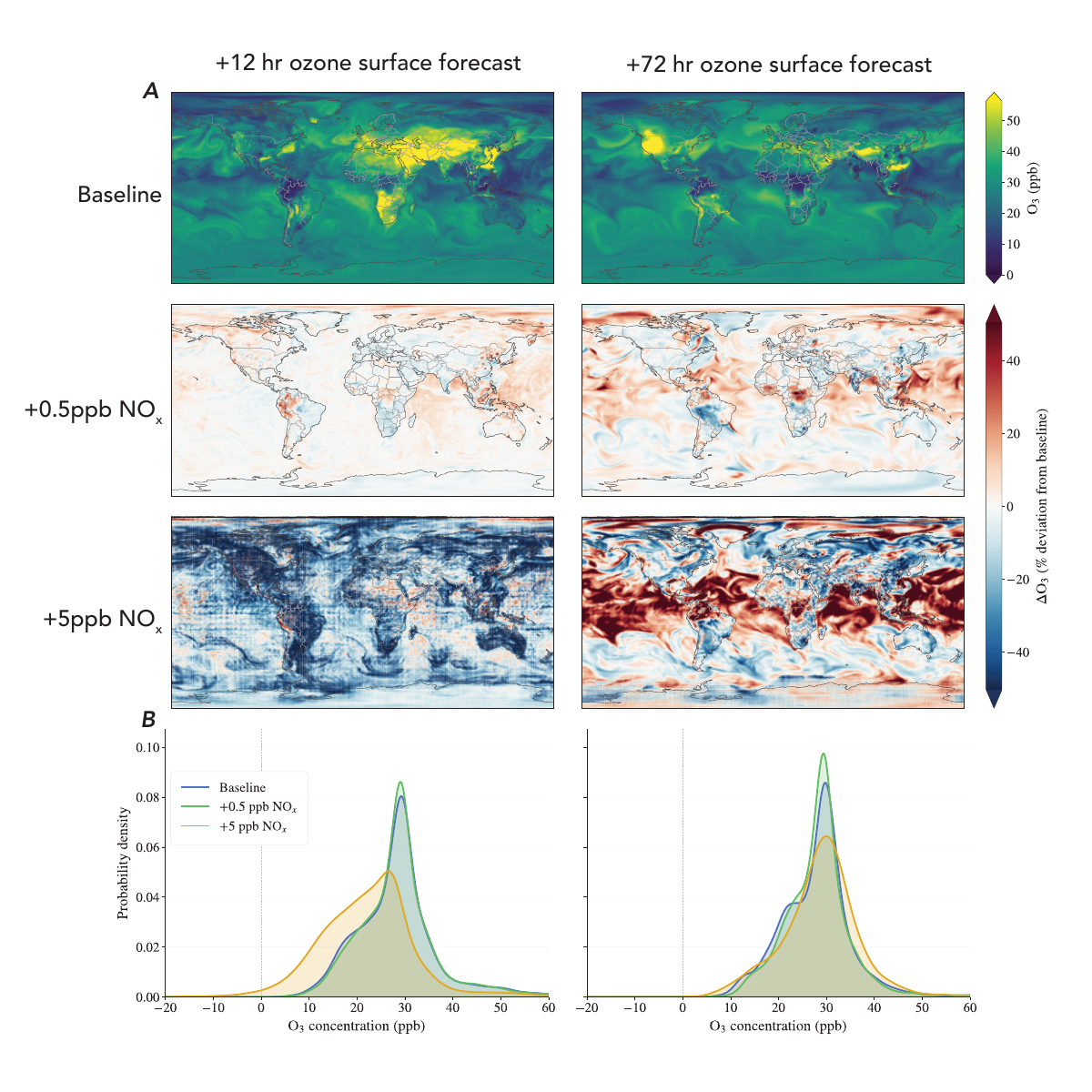}
    \caption{\textbf{Surface ozone response to global $\mathrm{NO}_x$ perturbations in Aurora.} Panel A shows the difference in surface ozone concentrations compared to a baseline where global $\mathrm{NO}_x$ is uniformly increased by 0.5 and 5 ppb for every grid cell at every atmospheric pressure level over both initialization timesteps. Panel B shows the probability distribution of surface ozone concentrations under the different baseline and elevated $\mathrm{NO}_x$ conditions after 12 and 72 h. After 12-h in the 5 ppb $\mathrm{NO}_x$ simulation, the surface ozone distribution contains negative concentrations. }
    \label{fig:nox_ablation}
\end{figure}

The smaller +0.5 ppb perturbation produces a weaker and more spatially heterogeneous response. Several tropical land regions, including parts of South America, central Africa, and Southeast Asia, show localized ozone increases rather than decreases at 12-h. This pattern is potentially consistent with ozone production regimes in which the ozone response to $\mathrm{NO}_x$ depends on the availability of $\mathrm{VOCs}$  and radicals. These tropical areas have high levels of biogenic $\mathrm{VOCs}$ and may therefore have sufficient oxidants to convert $\mathrm{NO}$ to $\mathrm{NO}_2$ without titrating $\mathrm{O}_3$. However, because Aurora does not include $\mathrm{VOCs}$ or radical species as prognostic variables, this result should be interpreted cautiously. The model may be using learned geographic, land-surface, meteorological, or climatological correlations that covary with chemical regime rather than representing $\mathrm{VOC}$ -sensitive chemistry directly.

By 72-h lead time, the perturbation response remains visible and becomes more spatially complex. This behavior is consistent with the broader finding of Bodnar et al. that Aurora Air Pollution performs best at short lead times and degrades during multi-step rollout. Our perturbation experiments extend this evaluation by tracking how a controlled chemical disturbance evolves through the learned dynamics. For the +5 ppb $\mathrm{NO}_x$ experiment, the median surface ozone concentration shifts from 6.1 ppb below baseline at 12-h to 0.5 ppb above baseline at 72-h. Thus, the perturbation not only persists but also changes sign in the global distribution. This persistence suggests that the imposed $\mathrm{NO}_x$ perturbation is not rapidly dissipated or chemically re-equilibrated in the learned dynamics. Instead, Aurora propagates a memory of the perturbation through the rollout, eventually producing regions of both ozone enhancement and suppression. It is not clear if Aurora, which may learn time-varying emission sources implicitly, is extending an emissions perturbation of $\mathrm{NO}_x$ or rolling out a response consistent with atmospheric chemical relationships. As the model conflates both emissions (a boundary condition) and chemistry (state-space evolution), it is not possible to further interpret its response to an initial condition perturbation. This behavior highlights a central limitation of end-to-end composition forecasting without explicit processes. That is, the model can learn plausible short-range sensitivities while still producing long-lived perturbation responses that are difficult to interpret as physical chemical adjustment (See Text S1, Figure S2).

\subsection{Wildfire Plume Persistence}

We use the July 2022 Courel-Valdeorras wildfire complex in Galicia, Spain, as a representative example of Aurora Air Pollution's response to emissions-driven pollution anomalies (Figure 4). The fires began on July 14 and burned approximately 25,600 hectares. In CAMS, total column $\mathrm{CO}$ remains elevated near the Galicia fire region and a visible plume is advected northward over the first 48-h of the forecast. In contrast, Aurora does not retain the localized wildfire-enhanced $\mathrm{CO}$ structure to the same degree as CAMS. By 24-h lead time, Aurora produces a smoother $\mathrm{CO}$ column with reduced enhancement near the fire detections and less distinct downwind plume structure. This difference persists at longer lead times and appears across the other two case studies (Figures S3, S4).

This behavior suggests that Aurora Air Pollution does not reliably convert fire-related composition anomalies in the initialized atmospheric state into sustained wildfire plume forecasts. Although the model is initialized from atmospheric states that contain the wildfire signal, it does not receive an explicit dynamic fire emissions flux or process-based emissions tendency during rollout. As a result, localized wildfire enhancements appear to be smoothed or relaxed toward broader background composition patterns. This limitation is potentially important for air quality applications because localized plume maxima (rather than broad regional averages) often drive exposure-relevant pollution episodes. 

\begin{figure}
    \centering
    \includegraphics[width=1.0\linewidth]{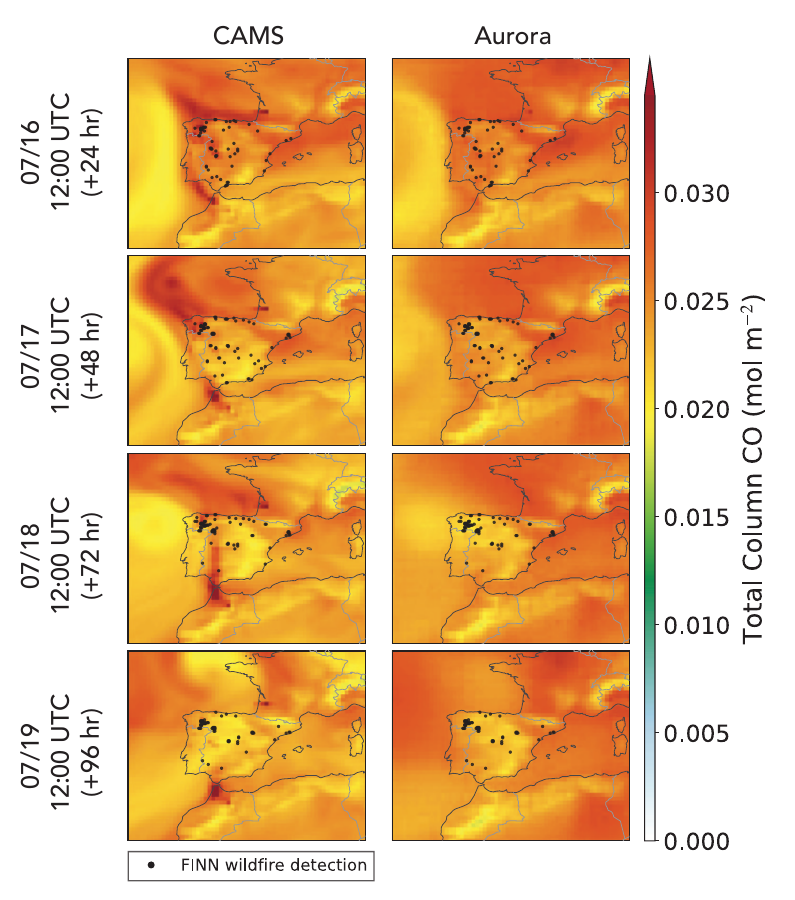}
    \caption{\textbf{Wildfire emissions and transport over Spain in CAMS and Aurora.} Both CAMS and Aurora Air Pollution are initialized on July 15th, 2022 +12:00 UTC. The black dots are fire detections across Spain according to the FINN v2.4 biomass burning emissions inventory \citep{gmd-16-3873-2023}. We plot total column CO, which is a typical tracer used to track biomass burning. }
    \label{fig:spain_wildfire}
\end{figure}

\newpage

\subsection{Spatial Embedding Structure}

The PCA visualizations show that Aurora Air Pollution's latent representations contain coherent spatial structure across both encoder and decoder states (Figure 5). The top three principal components explain approximately 45-60\% of the variance at each latent level. Thus, a small number of directions capture a large fraction of latent space variability. In Figure 5, the first three principal component scores are mapped to RGB channels to visualize spatial organization. The colors therefore do not correspond to specific physical or chemical variables; they only indicate coherent spatial variation within each independently-fitted PCA.

Across the encoder, these dominant modes are visually associated with broad geographic, topographic, and circulation-like structure. Latent level 0, which encodes surface variables, contains prominent land surface and orographic features, including clear signatures of the Andes and Tibetan Plateau. Higher encoder latent levels become progressively smoother and more spatially organized at large scales with patterns that resemble ascending vertical stages of  atmospheric circulation. This progression is consistent with the increasing vertical information encoded across latent levels, although PCA alone cannot assign a unique altitude range or physical variable to any individual component.

The decoder latent levels also contain coherent global structure, but their dominant patterns differ from the encoder in ways that may reflect some task-specific atmospheric chemistry fine-tuning. Decoder levels 0 and 3 retain broad meteorological organization similar to the analogous encoder levels, suggesting that much of the latent structure remains tied to the weather and transport inductive bias learned during pretraining. Decoder latent level 1 contains spatial structure that may resemble human activity or emissions-related patterns, and decoder latent level 2 introduces a sharp day-night boundary. Because photochemistry depends strongly on sunlight, the day-night boundary may reflect the added time-of-day variables or photochemically relevant diurnal variation. However, PCA alone cannot determine whether this mode encodes photolysis, radiation, or boundary-layer meteorology. Similarly, the apparent human activity pattern may reflect the static anthropogenic emissions fields or other correlated geographic information \citep{bodnarFoundationModelEarth2025}.

Overall, these PCA results suggest that the largest directions of variation in Aurora Air Pollution's latent spaces are dominated by meteorological and geographic structure with some decoder modes that may relate to atmospheric chemistry. This encoder-decoder contrast is consistent with a model that retains much of the pretrained meteorological representation while modifying downstream latent representations for air pollution forecasting. Similar structures appear in the monthly and annual PCA analyses (Figures S5, S6). These results do not imply that chemical information is absent from the model. Rather, the chemistry-specific structure is not cleanly separable from the dominant meteorological and geographic modes in the latent representations. This entanglement may limit the model's ability to learn features that do not depend on weather alone. This is not inherently a drawback of fine-tuning. Fine-tuning tends to preserve pretrained representations that support generalization \citep{kumarFineTuningCanDistort2022a}. The choice of fine-tuning regime can then determine how well a model adapts to a new task, such as atmospheric chemistry \citep{liRethinkingHyperparametersFinetuning2020, leeSurgicalFineTuningImproves2023}. 



\begin{figure}
    \centering
    \includegraphics[width=0.99\linewidth]{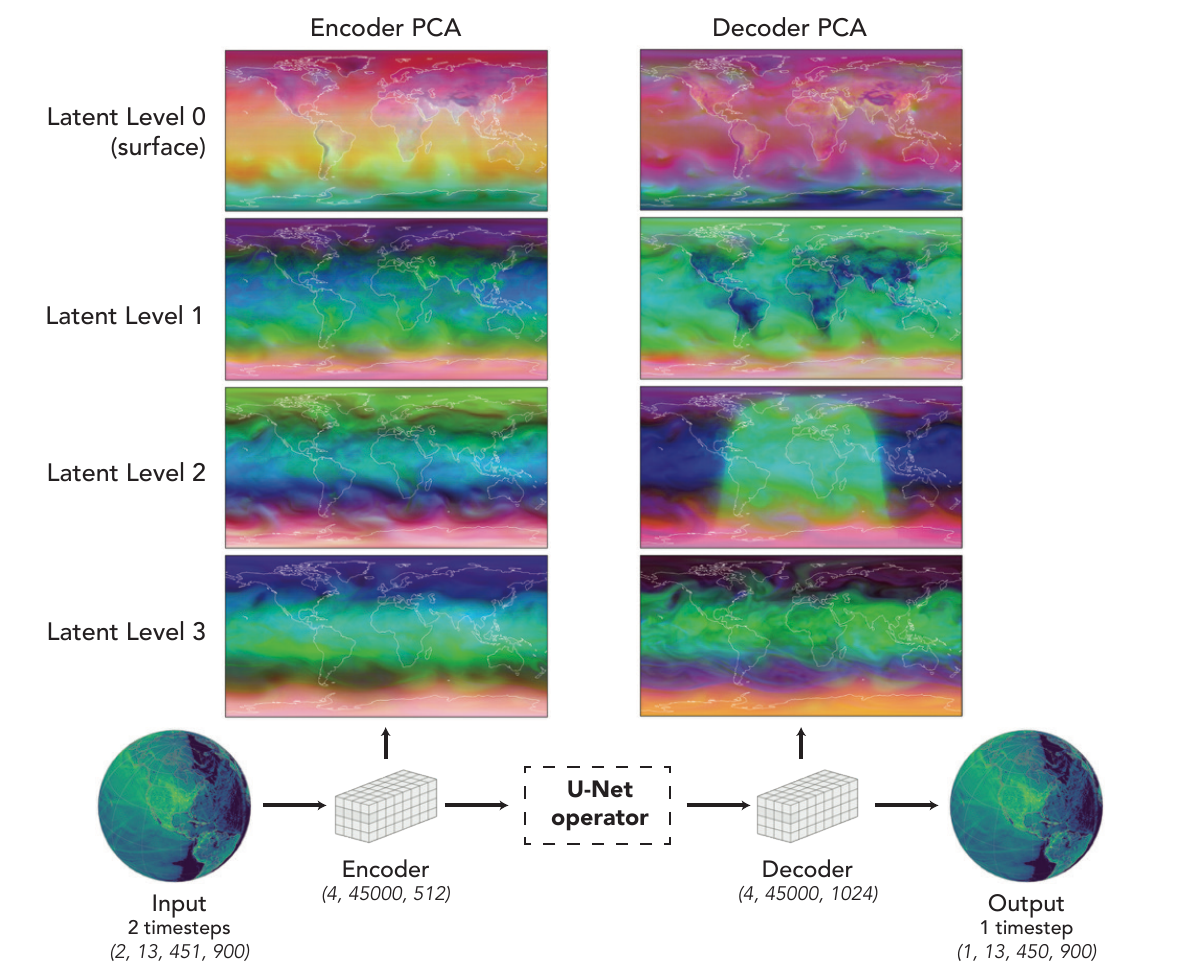}
    \caption{\textbf{PCA uncovers coherent structures in Aurora’s latent levels.} We perform PCA over the input latent and output latent levels of a single forward pass of Aurora Air Pollution initialized on September 1st, 2022 at 00:00 UTC. We map these 3 principal components to RGB space. The colorscale has no inherent meaning, but differences in colors suggest differences in the principal components.}
    \label{fig:pca}
\end{figure}

\subsection{Feature Discovery \& Causal Steering}

\textit{AuroraScope} SAEs reconstruct Aurora Air Pollution operator activations with high fidelity across most U-Net stages (Table 1). Reconstruction performance is highest in the early encoder and late decoder stages. In these stages, the mean fraction of variance explained exceeds 93\%. Performance is lower in the deepest encoder stage and earliest decoder stage, where mean fraction of variance explained is approximately 78 to 79\%. We hypothesize that this stage dependence reflects the spatial attention structure of the Aurora operator (Table S1). In the first two encoder stages and last two decoder stages, patches are constrained to attend to adjacent patches. In contrast, in the final encoder stage and first decoder stage, each patch can attend to all other patches. Without spatially-enforced structure, an SAE cannot reconstruct the features in these two stages as accurately. More distributed representations in these deeper stages may be less easily captured by a sparse feature basis \citep{cunninghamSparseAutoencodersFind2023}.



\begin{table}[t]
\caption{Fraction of variance explained (FVE) by \textit{AuroraScope} SAEs across U-Net operator stages.}
\label{tab:fve}
\centering
\begin{tabular}{llrcc}
\toprule
Component & Stage & Dimension & Mean FVE (\%) & Range (\%) \\
\midrule
Encoder & 0 & 512  & 99.7 & 99.6--99.8 \\
Encoder & 1 & 1024 & 97.4 & 95.2--99.9 \\
Encoder & 2 & 2048 & 77.8 & 75.8--80.4 \\
\addlinespace
Decoder & 0 & 2048 & 79.0 & 76.4--82.3 \\
Decoder & 1 & 1024 & 93.7 & 91.2--96.2 \\
Decoder & 2 & 512  & 99.7 & 99.6--99.7 \\
\bottomrule
\end{tabular}
\end{table}
The learned \textit{AuroraScope} features reveal multiple forms of spatial organization inside the Aurora Air Pollution operator (Figure 6). Some features activate over coherent regional structures, such as plume-like or circulation-aligned patterns over ocean basins and continental outflow regions. Other features activate over broad portions of the globe, suggesting globally coherent atmospheric state information rather than localized chemistry. A third class of features follows wave-like or day-night patterns, which may reflect transport, radiative forcing, or temporal structure in the forecast state. We also identify features that appear to be artifacts of the model grid or architecture. For example, one high activation feature is concentrated almost entirely along a single latitude band indicating that Aurora uses at least some spatially artificial structures internally when producing forecasts. These examples show that SAEs can expose organized internal representations in Aurora, but high activation or high variance contribution does not imply that a feature corresponds to a chemically plausible mechanism. Additional SAE features examples are found in the Supplementary Information (Figure S7). 

\begin{figure}
    \centering
    \includegraphics[width=1\linewidth]{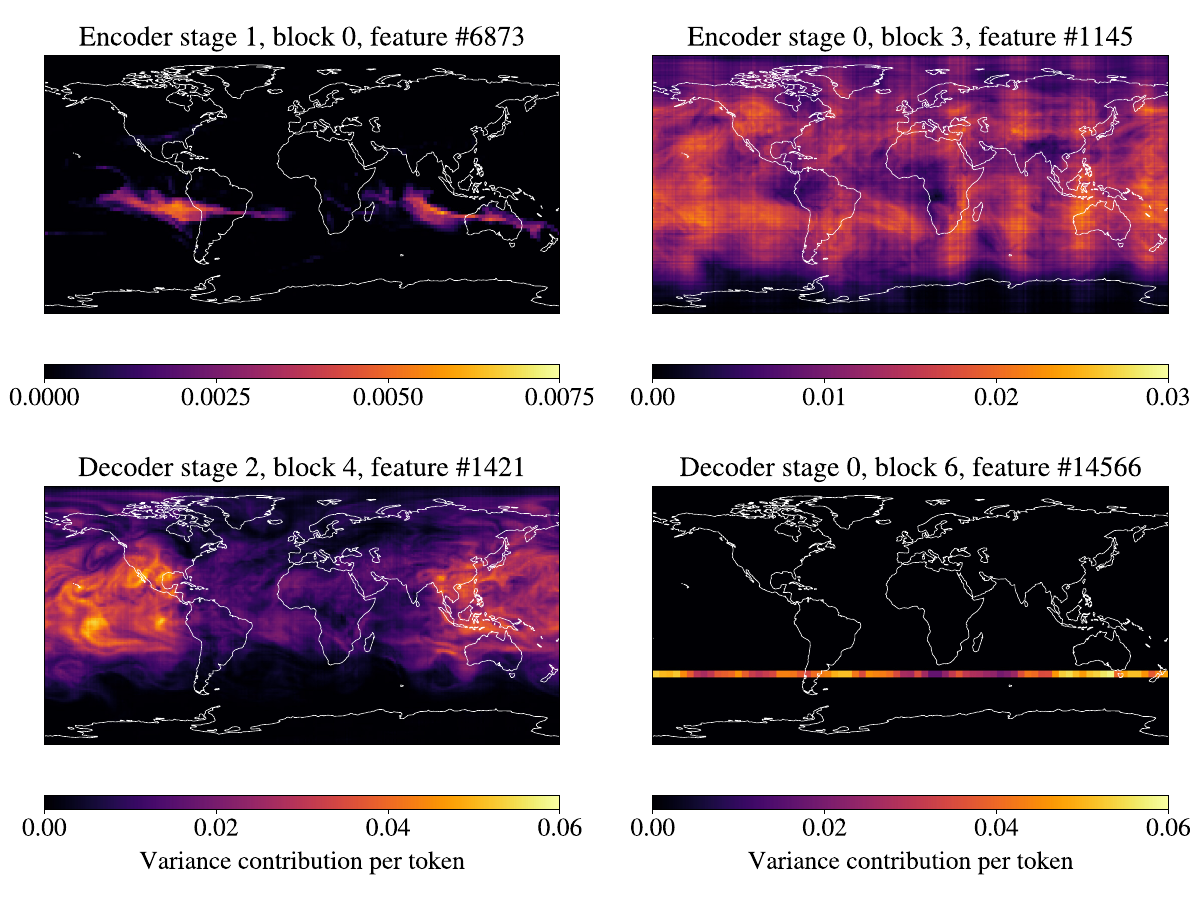}
    \caption{\textbf{Example features learned by \textit{Aurorascope} SAEs. }A range of features are represented, including regional transport (upper left), whole-globe activation (upper right), diurnal cycles (lower left), and a spurious feature for a latitude band (lower right). All features are extracted from the SAEs on the first step of a rollout initialized at September 01, 2022 +00:00 UTC. Resolution differences are reflective of downsampling dependent on the U-Net stage.}
    \label{fig:sae_features}
\end{figure}

Steering experiments test whether these features causally influence chemical forecasts. We zero-ablate the top ten features across  \textit{AuroraScope}  ranked by maximum per-token variance contribution and measure the resulting global mean absolute change in surface ozone and $\mathrm{NO}_2$ after one 12-h forecast step. We present two of these features as case studies (Figure 7). The first feature (\#2712) is the highest ranked overall and produces the largest change in both species, shifting global mean absolute surface ozone by 0.84 ppb and $\mathrm{NO}_2$ by 0.063 ppb. The second feature (\#2866) produces the second largest change in $\mathrm{NO}_2$ and changes surface ozone by 0.82 ppb and $\mathrm{NO}_2$ by 0.037 ppb. Although these global means are small, individual grid cell changes can be large. For example, the first feature changes surface ozone by up to +24 and -16 ppb and $\mathrm{NO}_2$ by up to +57 and -19 ppb in single grids. The ablation response maps show structured changes rather than spatially random perturbations. Removing a single \textit{AuroraScope} feature produces contiguous regions of increased or decreased ozone and $\mathrm{NO}_2$, often following broad geographic or meteorological structure. These responses show that the selected features are causally involved in the forecast rather than being descriptive summaries of the residual stream.

However, the induced changes are not cleanly species-specific, process-specific, and are not straightforwardly interpretable as atmospheric chemistry. Ablating either feature modifies ozone, $\mathrm{NO}_2$, and the other pollutants simultaneously. For feature \#2712, the ablation produces sharp land-sea gradients in ozone and alternating positive and negative changes over the ocean that appear tied to large-scale circulation rather than to ozone production or titration regimes.  For feature \#2866, the $\mathrm{NO}_2$ response includes localized changes near apparent emissions hotspots, but these do not consistently align with known population centers or major emitting regions. This $\mathrm{NO}_2$ response also contains adjacent positive and negative changes over small spatial scales, which is difficult to interpret as a physically driven emissions response. Because Aurora does not explicitly represent time-varying emissions, transport, or chemical mechanisms, it is difficult to assign these steered responses to specific atmospheric processes.


\begin{figure}
    \centering
    \includegraphics[width=1\linewidth]{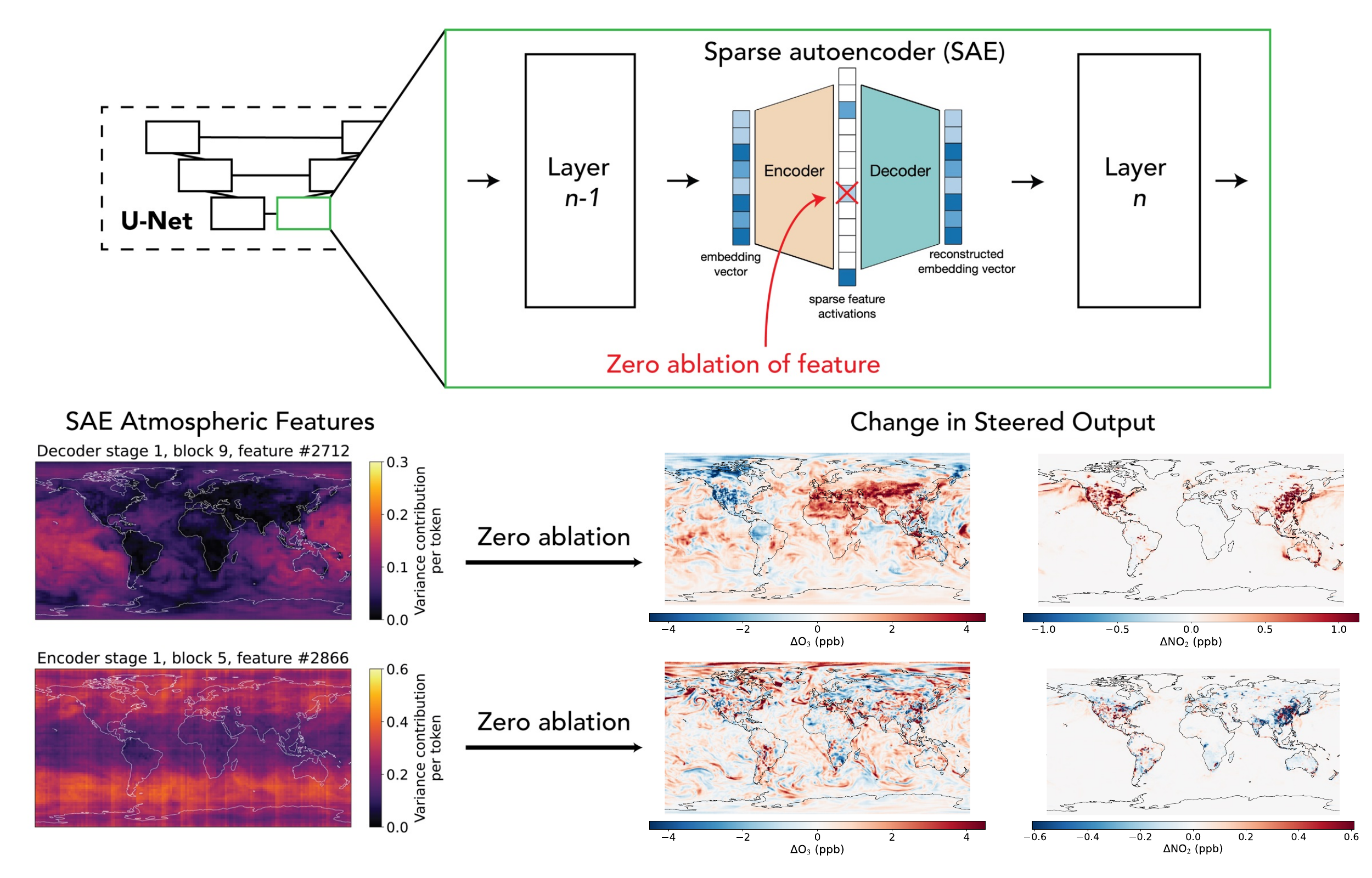}
    \caption{\textbf{Steering Aurora forecasts by zero-ablating \textit{AuroraScope} features.} Removing individual feature contributions from the operator residual stream changes the resulting 12-h ozone and $\mathrm{NO}_2$ forecasts which shows that the features causally influence chemical output. The induced changes are spatially structured but chemically entangled.}
    \label{fig:sae_steering}
\end{figure}
These steering results illustrate a broader challenge for mechanistic interpretability of atmospheric chemistry FMs. Aurora Air Pollution already exhibits nonphysical behavior in several output diagnostics, including negative concentrations, patch-scale artifacts, and weak retention of emissions plumes. Interpreting internal features in such a model is therefore more difficult than assigning physical meaning to features in a model that already satisfies basic chemical constraints. Nonetheless, this difficulty is evidence in favor of the usefulness of mechanistic interpretability. As AI forecasting systems are increasingly used under nonstationary conditions, including changing emissions landscapes, increasing wildfire activity, and climate-driven shifts in meteorology, forecast skill on historical reanalysis data may not be sufficient to establish reliability.

\section{Conclusion}

We present the first mechanistic interpretability analysis of an AI foundation model fine-tuned for atmospheric chemistry. We pair chemically motivated tests of Aurora Air Pollution's forecasts with direct analysis of its internal representations. This combined analysis shows that skillful air pollution prediction can coexist with internal organization governed more by inherited meteorology and architectural artifacts than by chemistry. This organization has direct consequences for forecasting. It smooths emission features and raises the risk of poor generalization under the out-of-distribution conditions most relevant to air quality forecasting. Aurora reproduces broad composition patterns at short lead time, but it does so without the mechanistic and positivity constraints that a process-based model enforces. Even after chemistry fine-tuning, the dominant modes of its latent representations remain strongly aligned with the weather and transport inductive bias established during pretraining. However, we do observe some later decoder modes that may carry more chemistry-related information. We cannot determine from these experiments why the meteorological organization persists so strongly. Plausible contributors include the low-rank nature of the fine-tuning, which updates only a small subspace of the pretrained weights (i.e., LoRA) and the general difficulty of displacing representations acquired during pretraining.

Aurora Air Pollution's chemical behavior is mixed. It recovers an ozone-suppression response to elevated reactive nitrogen concentrations, reproducing the $\mathrm{NO}$ titration expected in $\mathrm{NO}_x$-rich air and showing that some coupled chemistry is present in the learned dynamics. Whether it has learned the nonlinear, regime-dependent characteristics of that chemistry is more difficult to establish given the limited set of five prognostic trace gases and the absence of process-level outputs. Small nitrogen perturbations raise ozone in a few tropical regions, which may suggest $\mathrm{VOC}$-saturated production. However, because Aurora carries neither $\mathrm{VOCs}$ nor radicals as outputs, it may instead be tracking geography or meteorology that covaries with chemical regime rather than the chemistry itself. At the same time, the model drives component concentrations negative over a large fraction of grid cells, loses photochemical consistency as the rollout proceeds, and relaxes sharp emissions features such as wildfire plumes toward background. Aurora Air Pollution has therefore learned useful short-range chemical sensitivities without acquiring the constraints that keep a process-based model physically consistent.

Our model internal analyses show both the promise and the current limits of this approach. Using \textit{AuroraScope}, we recover interpretable features corresponding to global transport, regional composition differences, and diurnal structure. We show through zero ablation that these features causally influence the chemical forecast with the highest-ranked feature changing global mean ozone by $\sim$0.84 ppb and $\mathrm{NO}_2$ by $\sim$0.06 ppb, while changing individual surface grid cells by as much as 24 ppb for ozone and 57 ppb for $\mathrm{NO}_2$. Yet the induced responses are entangled across species and are difficult to attribute to specific atmospheric processes, in part because Aurora Air Pollution has no physically well-behaved forecast against which to interpret them. More robust attribution will require testing learned features against independent reference fields for known processes, such as chemical production and loss rates. The principal questions are whether these process quantities are recoverable from the activations (e.g., linear probes) and whether a feature activates preferentially when the corresponding process dominates. These tests will become far more informative when applied to models that respect positivity and preserve emissions structure. The same holds for tracking how features emerge or persist across the transition from the pretrained to the fine-tuned model. Extending our full interpretability workflow across fine-tuning checkpoints and across different Earth system FMs is a natural next step.

The need for evaluating Earth system FMs will only grow. Once an Earth system FM has been pretrained, it can be fine-tuned for a new prediction task at a small fraction of the cost of building a conventional CTM or Earth system model. This efficiency is expected to make fine-tuned FMs a widespread tool for atmospheric composition prediction \citep{hickmanApplicationsMachineLearning2025a}. The fine-tuning procedure that produced Aurora Air Pollution can be applied by operational forecasting centers, environmental agencies, and individual research groups to a wide range of composition targets. As these models move into air quality forecasting and other settings that inform public health and regulatory decisions, skill on historical reanalysis alone will not establish reliability. These FMs were never trained on the shifting emissions, intensifying wildfire activity, and changing meteorology that a warming climate brings. Our results also raise a more basic question. It remains unclear whether an FM carrying such a strong meteorological inductive bias is well suited to a target like tropospheric ozone, whose production is nonlinear and depends on emissions, photochemistry, and deposition rather than on weather alone \citep{kelpAI4O3FoundationalData2025a}. Establishing standards for the physical evaluation and governance of atmospheric chemistry FMs will be an essential foundation as these models take on a larger role in Earth system science. Internal mechanisms and chemical understanding, rather than benchmark scores alone, should determine when an AI forecast is trustworthy.

\section*{Open Research Section}
The trained SAEs, metadata index, and standalone loading code are publicly released as \textit{AuroraScope} at \url{huggingface.co/hujason/aurorascope}. Code to reproduce figures and tables will be made available upon publication.

\section*{Conflict of Interest declaration}
The authors declare there are no conflicts of interest for this manuscript.

\acknowledgments
The authors thank the Stanford Research Computing Center for providing computational resources and support. This research was supported by the NOAA Climate and Global Change Postdoctoral Fellowship Program, administered by UCAR's Cooperative Programs for the Advancement of Earth System Science (CPAESS) under the NOAA Science Collaboration Program award \# NA23OAR4310383B. M.K. acknowledges support from Stanford University. The authors thank Qindan Zhu and James Weber for valuable discussions, and Marshall Burke for undergraduate research sponsorship. I.H-M thanks the Stanford Data Science Fellowship for its support.

%
%
\newpage
 \bibliography{agusample} 

%
%
\clearpage

\noindent\textbf{Text S1.}
 To test whether a localized emission perturbation propagates globally through Aurora Air Pollution's learned dynamics, we perturb a single grid cell column rather than the global field used in the main text experiments. Beginning from the September 1, 2022 00:00 UTC initialization, we add 100 ppb of $\mathrm{NO}_x$ in the grid cell nearest Beijing (39.9°N, 116.4°E) across all 13 pressure levels and both input timesteps. All other cells are left at baseline values. We run a five-day rollout and compare it against the unperturbed forecast and evaluate the surface (1000 hPa) changes in ozone and $\mathrm{NO}_2$ at 12- and 72-h lead times (Figure S2). The response is spatially incoherent and affects many implausible locations. At 12-h lead time, the perturbation shifts ozone by up to 2.1 ppb near the source cell, although the global mean change is only 0.0006 ppb. We observe plausible ozone increases over Japan, but puzzling ozone decreases over India and southern Russia. For $\mathrm{NO}_2$, we observe large-scale changes across East Asia and India, which is unrealistic given that $\mathrm{NO}_2$ perturbations should remain close to their point source. By 72-h lead time, the perturbation spreads farther rather than dissipating. The ozone response extends across essentially all surface fields (although the concentrations are small enough that this is most likely numerical noise), while the $\mathrm{NO}_2$ response continues to grow and spread into the Middle East. Rather than remaining confined near the source or propagating as coherent transport would predict, the single column perturbation produces a diffuse and physically implausible pattern of far-field changes.

\clearpage
\setcounter{figure}{0}
\renewcommand{\thefigure}{S\arabic{figure}}
\setcounter{table}{0}
\renewcommand{\thetable}{S\arabic{table}}

\begin{figure}
    \centering
    \includegraphics[width=1\linewidth]{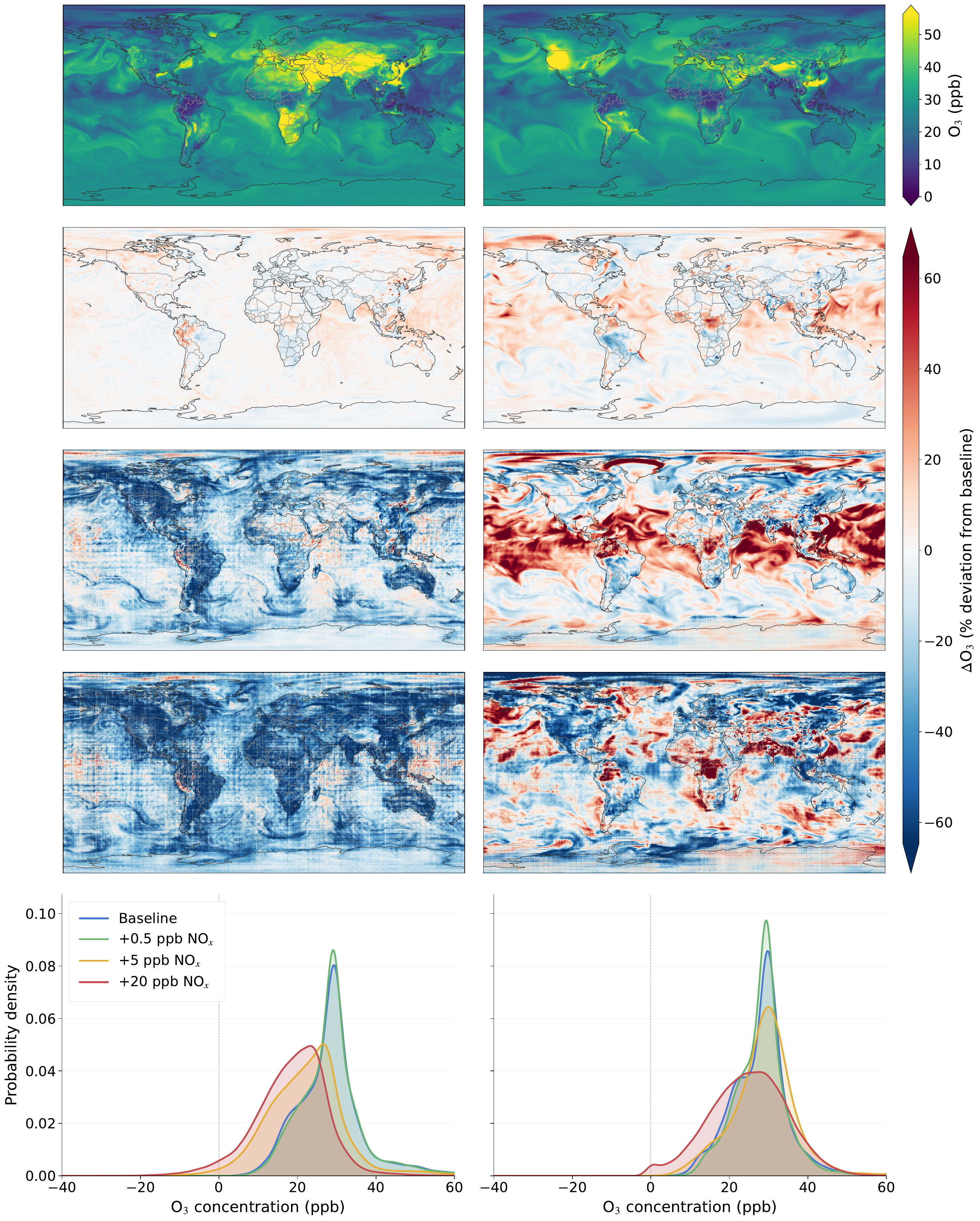}
    \caption{Surface ozone response to global  $\mathrm{NO}_x$ perturbations in Aurora as in Figure 3, but including +20 ppb experiment.}
    \label{fig:20ppb_fig}
\end{figure}

\begin{figure}
    \centering
    \includegraphics[width=1\linewidth]{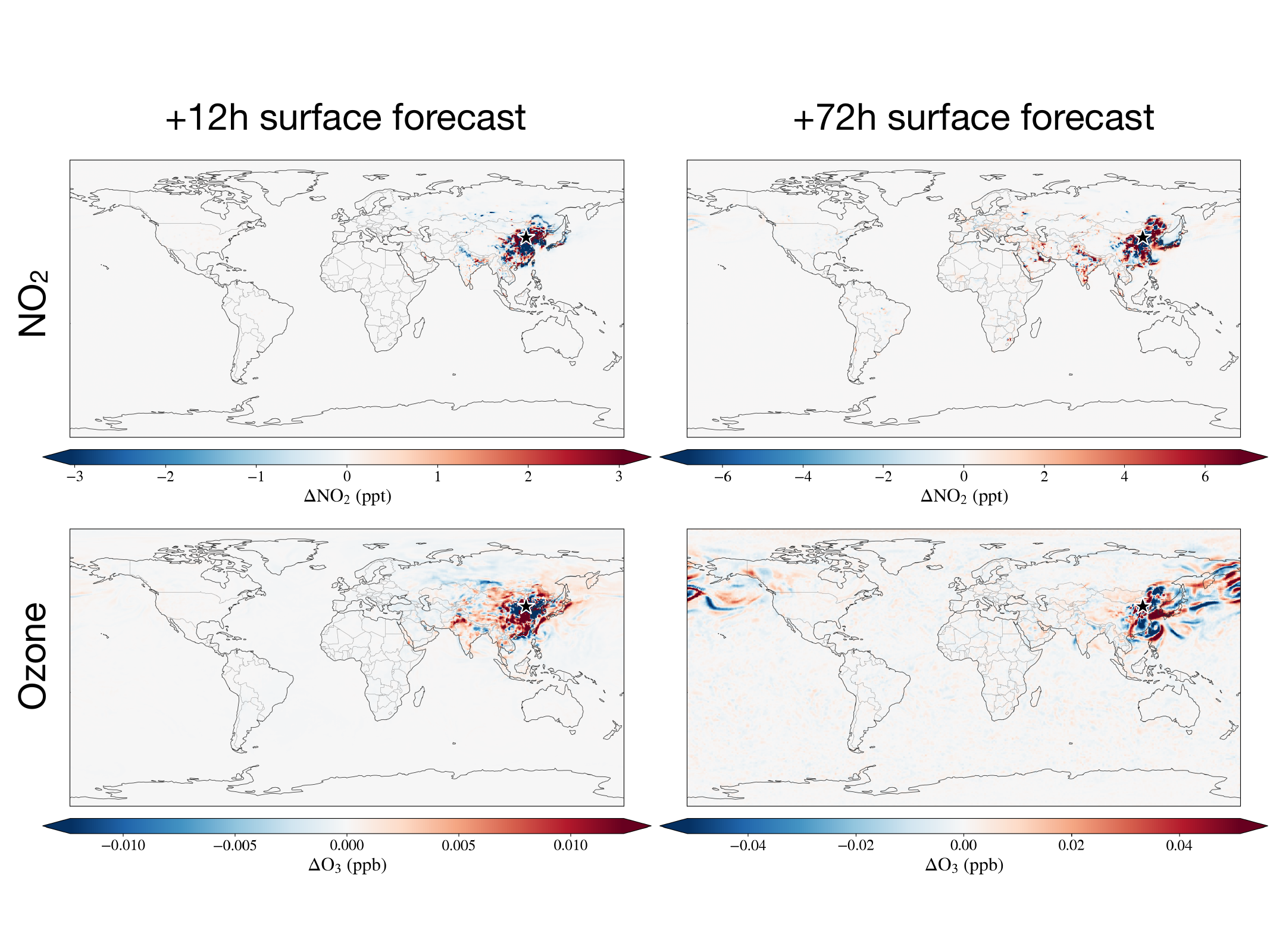}
    \caption{Surface ozone response to single column $\mathrm{NO}_x$ perturbations in Aurora. Panels shows the difference in surface ozone and $\mathrm{NO}_2$ concentrations compared to a control baseline. Location of perturbation indicated by star.}
    \label{fig:noxcolumn_fig}
\end{figure}

\begin{figure}
    \centering
    \includegraphics[width=1\linewidth]{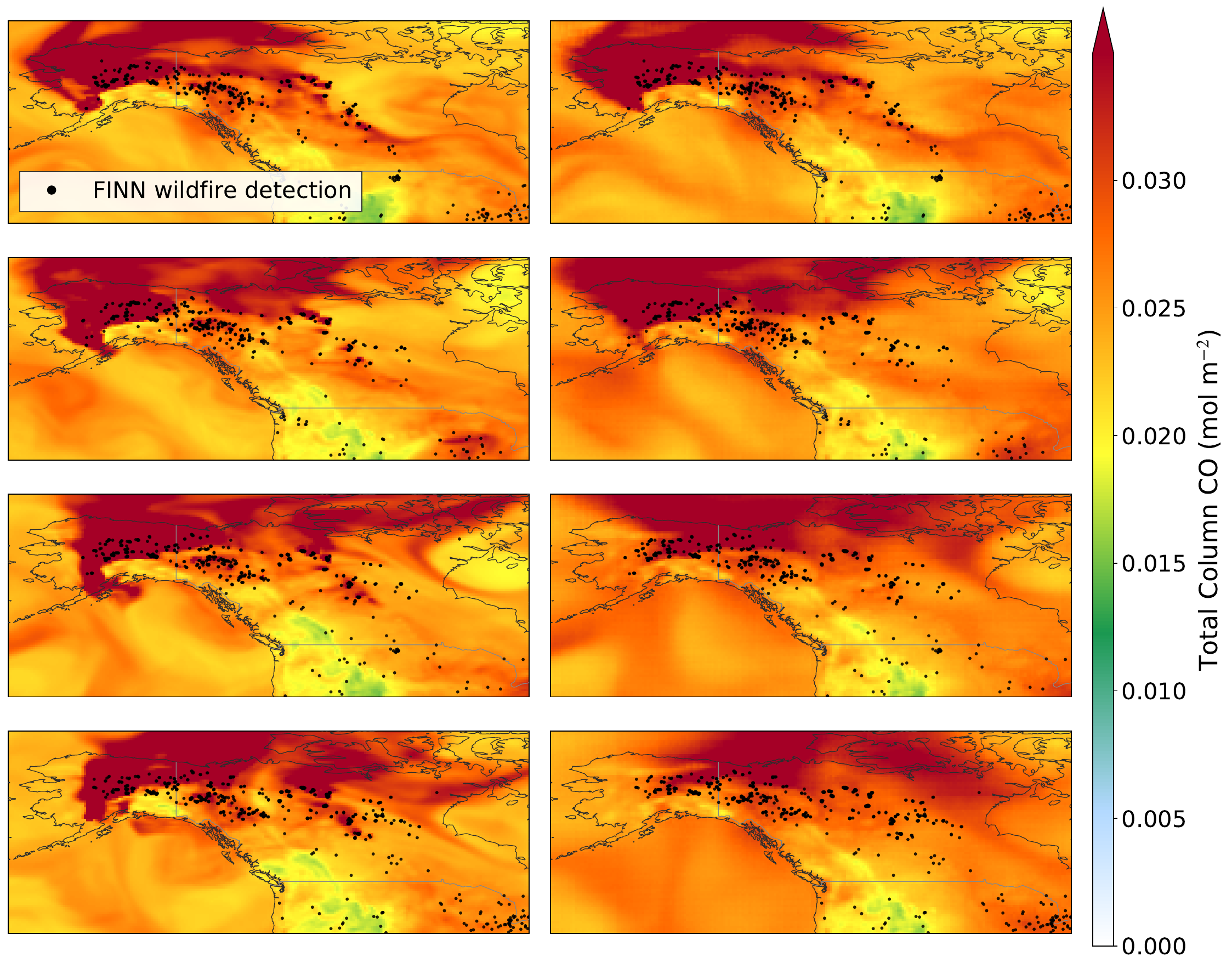}
    \caption{Wildfire emissions and transport in CAMS and Aurora as in Figure 4, but over Alaska. }
    \label{fig:alaska_wildfire}
\end{figure}

\begin{figure}
    \centering
    \includegraphics[width=1\linewidth]{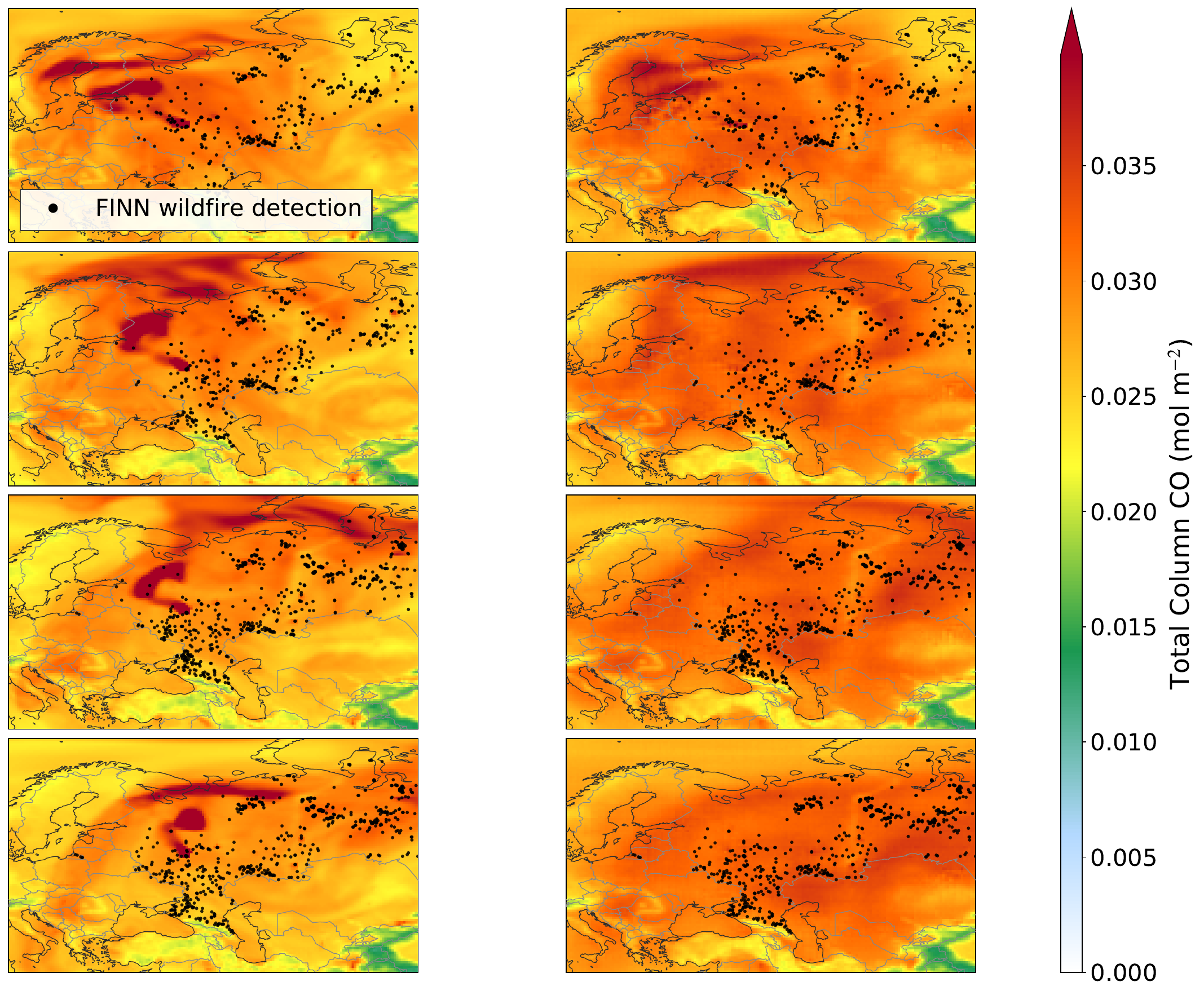}
    \caption{Wildfire emissions and transport in CAMS and Aurora as in Figure 4, but over central Russia. }
    \label{fig:russia_wildfire}
\end{figure}

\begin{figure}
    \centering
    \includegraphics[width=1\linewidth]{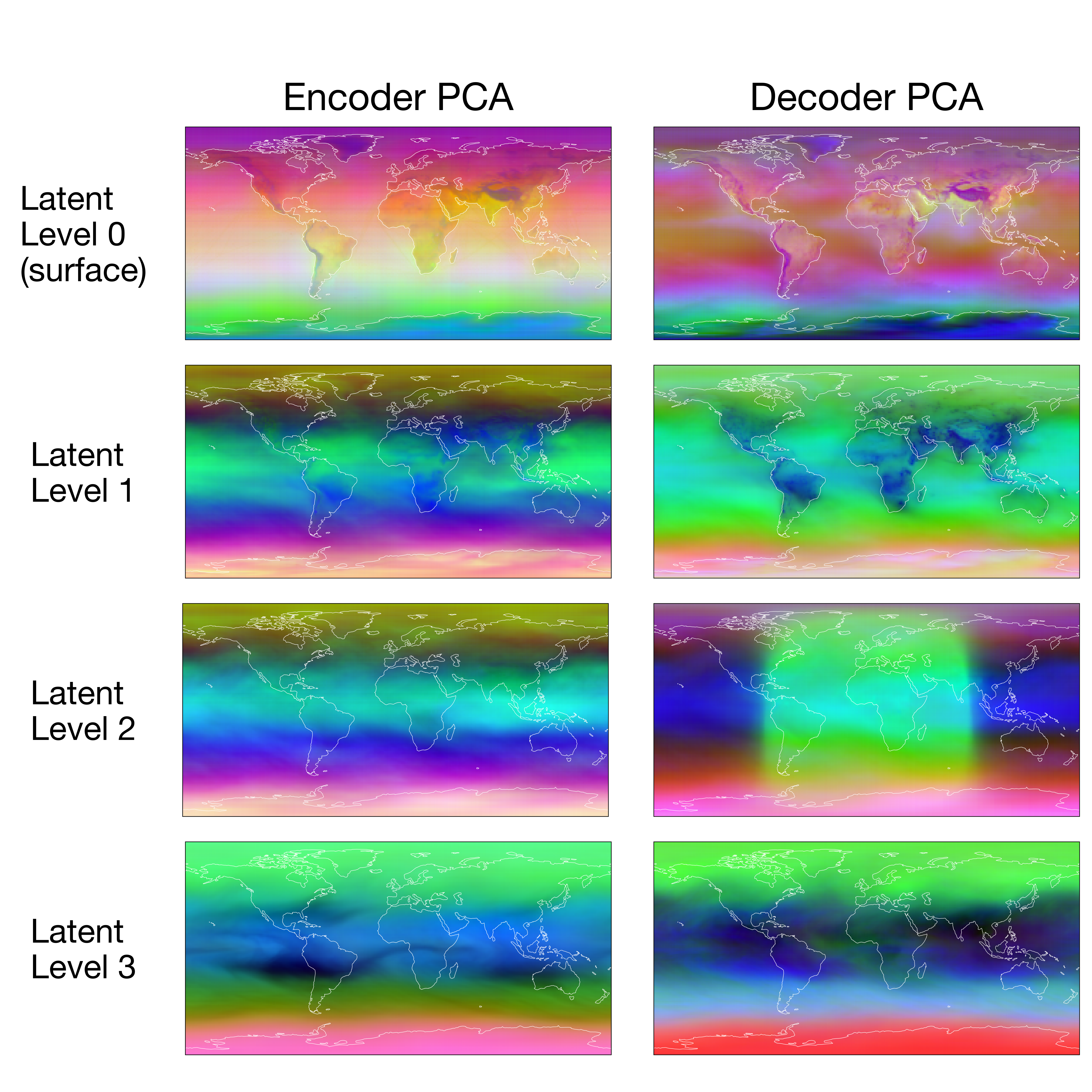}
    \caption{PCA over the input latent and output latent levels of a month-long (September, 2022) pass of Aurora Air Pollution. Only 00:00 UTC time steps are analyzed.}
    \label{fig:PCA_month}
\end{figure}

\begin{figure}
    \centering
    \includegraphics[width=1\linewidth]{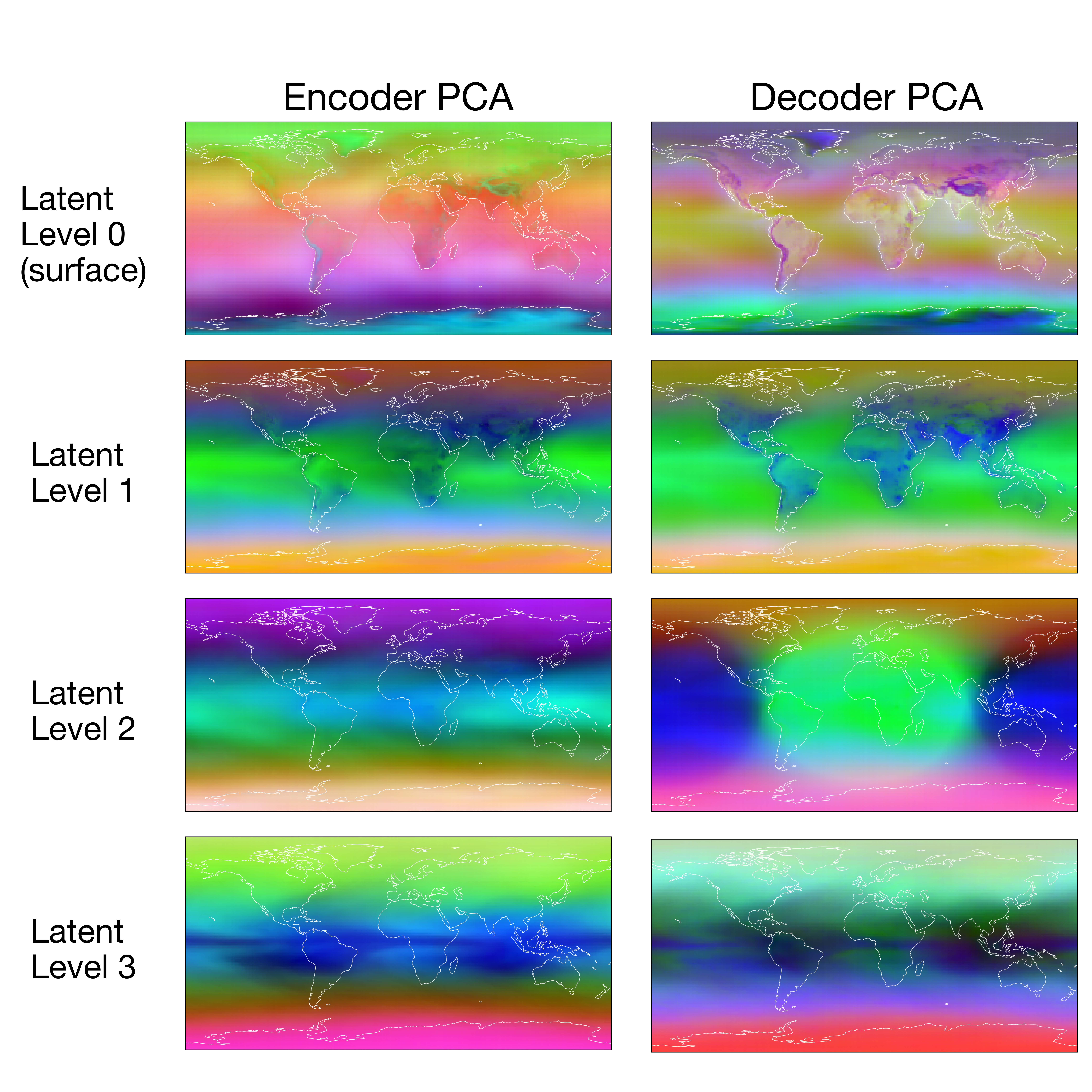}
    \caption{PCA over the input latent and output latent levels of a year-long (June 1, 2022 to May 31, 2023) pass of Aurora Air Pollution. Only 00:00 UTC time steps are analyzed.}
    \label{fig:PCA_year}
\end{figure}

\begin{figure}
    \centering
    \includegraphics[width=.92\linewidth]{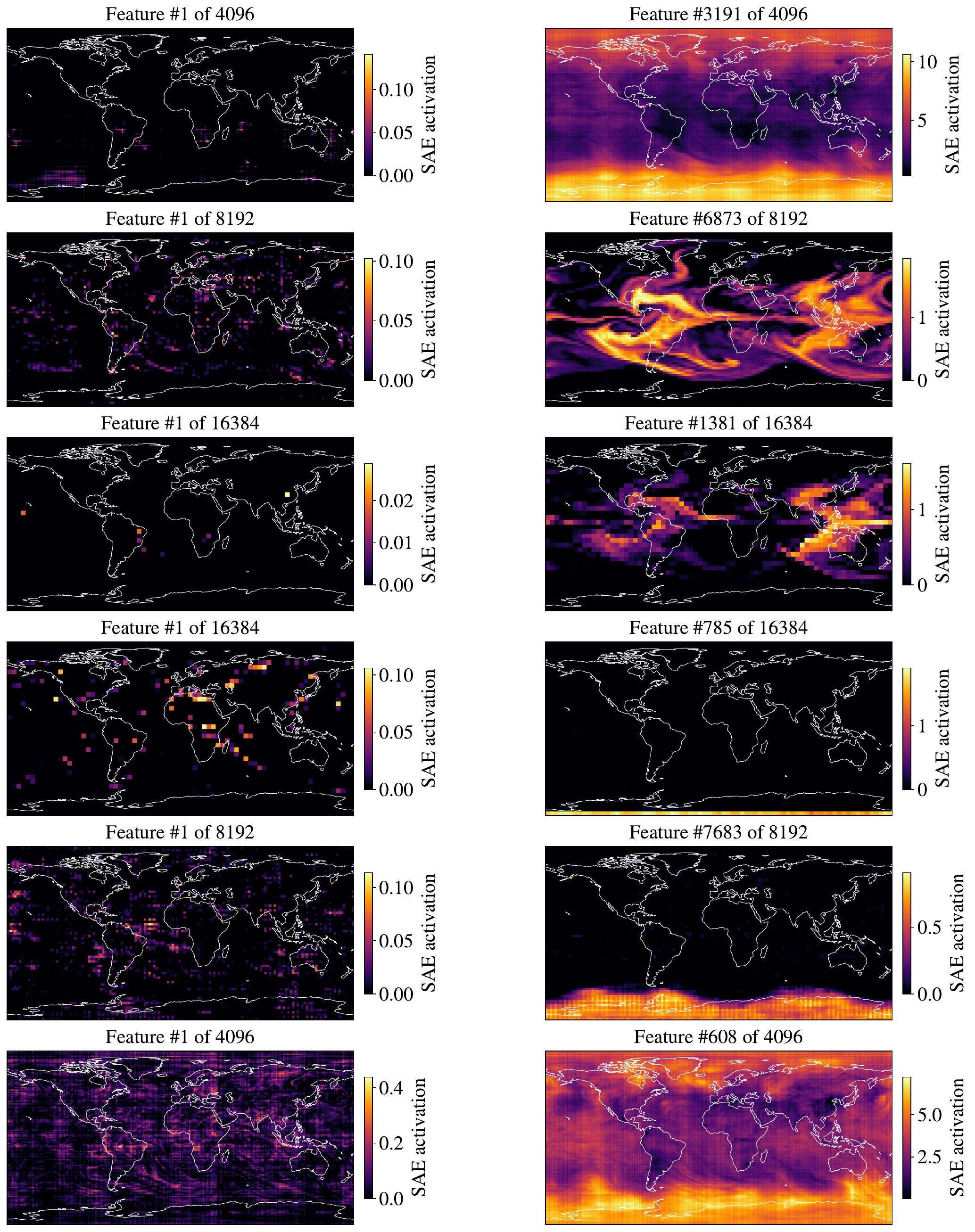}
    \caption{Additional example features learned by \textit{Aurorascope} SAEs as in Figure 6. The first column is a visualization of the first feature in each component of the U-Net operator stage (see Table S1). The second column contains illustrative examples of features that correspond to recognizable atmospheric patterns and examples of potentially spurious patterns.}
    \label{fig:sae_examples}
\end{figure}

\begin{table}[t]
\caption{Number of learned features (dictionary size $= 8\times d_{\mathrm{model}}$) per \textit{AuroraScope} SAE by U-Net operator stage.}
\label{tab:feature_counts}
\centering
\begin{tabular}{|l|l|r|r|r|}
\hline
Component & Stage & $d_{\mathrm{model}}$ & \# SAEs (blocks) & Features per SAE \\
\hline
Encoder & 0 & 512 & 6 & 4,096 \\
Encoder & 1 & 1024 & 10 & 8,192 \\
Encoder & 2 & 2048 & 8 & 16,384 \\
\hline
Decoder & 0 & 2048 & 8 & 16,384 \\
Decoder & 1 & 1024 & 10 & 8,192 \\
Decoder & 2 & 512 & 6 & 4,096 \\
\hline
\multicolumn{4}{|l|}{Total across all 48 SAEs} & 475,136 \\
\hline
\end{tabular}
\end{table}

\end{document}